\documentclass[lettersize,journal]{IEEEtran}
\usepackage{amsmath,amsfonts}
\usepackage{algorithmic}
\usepackage{algorithm}
\usepackage{array}
\usepackage[caption=false,font=normalsize,labelfont=sf,textfont=sf]{subfig}

\usepackage{relsize}
\usepackage{textcomp}
\usepackage{stfloats}
\usepackage{url}
\usepackage{verbatim}
\usepackage{graphicx}
\usepackage{cite}
\usepackage{xcolor}
\usepackage{ paralist}
\usepackage{stfloats}
\usepackage{tabularx}
\usepackage{bm}
\usepackage[numbers]{natbib}
\usepackage[tableposition=bottom]{caption}

\usepackage{booktabs}

\title{Sparsity exploitation via discovering graphical models in multi-variate time-series forecasting}
\author{
Ngoc-Dung Do \IEEEauthorrefmark{1} \IEEEauthorrefmark{3},
Truong Son Hy \IEEEauthorrefmark{2} \IEEEauthorrefmark{3} \IEEEauthorrefmark{4},
Duy Khuong Nguyen \IEEEauthorrefmark{1}
\thanks{\IEEEauthorrefmark{1} FPT Software AI Center, Hanoi 10000, Vietnam}
\thanks{\IEEEauthorrefmark{2} University of California San Diego, La Jolla, CA 92093, USA}
\thanks{\IEEEauthorrefmark{3} Co-first authors}
\thanks{\IEEEauthorrefmark{4} Correspondence to tshy@ucsd.edu}
}
\date{April 2023}

\begin{document}
\maketitle

\begin{abstract}

Graph neural networks (GNNs) have been widely applied in multi-variate time-series forecasting (MTSF) tasks because of their capability in capturing the correlations among different time-series. These graph-based learning approaches improve the forecasting performance by discovering and understanding the underlying graph structures, which represent the data correlation. When the explicit prior graph structures are not available, most existing works cannot guarantee the sparsity of the generated graphs that make the overall model computational expensive and less interpretable. In this work, we propose a decoupled training method, which includes a graph generating module and a GNNs forecasting module. First, we use Graphical Lasso (or GraphLASSO) to directly exploit the sparsity pattern from data to build graph structures in both static and time-varying cases. Second, we fit these graph structures and the input data into a Graph Convolutional Recurrent Network (GCRN) to train a forecasting model. The experimental results on three real-world datasets show that our novel approach has competitive performance against existing state-of-the-art forecasting algorithms while providing sparse, meaningful and explainable graph structures and reducing training time by approximately $40\%$. Our PyTorch implementation is publicly available at \url{https://github.com/HySonLab/GraphLASSO}.
\end{abstract}

\begin{IEEEkeywords}
Graph Neural Networks, sparse graph learning, graphical lasso, multi-variate time-series forecasting, spatio-temporal modeling.
\end{IEEEkeywords}

\section{Introduction}

Many real world problems can be formed in Multi-variate Time Series Forecasting (MTS) including traffic load forecasting \cite{GCRNtraffic, Dcrnn}, epidemic modeling \cite{pmlr-v184-hy22a}, 
retail sales \cite{retailsales}, finance \cite{finance}, etc. that involve recorded temporal data from multiple sensors being processed and aligned into the same time axis. 
Thus, MTS data can be seen as a type of spatial-temporal data, which contains multiple inter-related time series. A forecasting model needs to capture both spatial and temporal information in order to perform well in predicting for future values.

Previous work on MTS forecasting can be divided into statistics-based and deep learning-based approaches. The former often consider the linear correlations among variables (i.e. time series) \cite{VAR, VARMA}. Regarding the latter, early works have utilized Convolution Neural Networks (CNN) to capture the correlations among variables, yet ignored their non-Euclidean pairwise dependencies. Transformer-based approaches consist of spatial attention mechanism to discover the reciprocal salience of different spatial
locations at each layer \cite{logformer, crossformer}. Recently, Spatial-Temporal Graph Neural Networks (STGNNs) \cite{STGCN, STGNN, Dcrnn}, have attracted an increasing attention because these methods not only archive outstanding performance on forecasting tasks but are also partially interpretable based on the graph structures, \citep{AGRCNbai2020, li2017diffusion, MTGNN2020}. In general, STGNNs combine Graph Convolutional Networks (GCN) \cite{4700287, kipf2017semisupervised} to explore the relationship between time series and sequential models to capture the temporal dependencies. In this case, each time series is one node of the graph and the relationship between two time series is represented by an edge. In practice, these graphs need to be learnt from observations rather than to be pre-defined, and to be sparse in number of connections.


In the deep learning setting, end-to-end training methods often incorporate a graph learning module with a forecasting module to maximize performance on the downstream tasks, for example, GTS \cite{shang2021discrete} and  MTGNN \cite{MTGNN2020}. This approach is particularly sensitive with hyper-parameters and cannot guarantee the sparse property. Beside, end-to-end training may lead to difficulty in convergence and usually generates dense graphs. In short, most of the aforementioned graph-learning methods cannot ensure the sparsity, which is an importance constraint in real world problems.

In this paper, we propose a simple but reliable approach that can efficiently capture the graph structures as well as guarantee the sparsity of their adjacency matrices by using GraphLASSO \cite{10.1093/biostatistics/kxm045, 10.1093/biomet/asm018}. 
GraphLASSO is a sparse penalized maximum likelihood estimator for the precision matrix of a multivariate Gaussian distribution, which encodes conditional correlations between pairs of variables given the remaining variables. It not only provides an effective way to control the sparsity of the relationships but also reduces the training time by rapidly generating the graphs representing correlations without end-to-end training.
In summary, our work has three main contributions as following:
\begin{compactitem}
\item A method of decoupled training that has a separated process for graph generation via GraphLASSO,
\item Constructing graph structures and exploiting the spatial features of MTS for both static and times-varying cases,
\item Extensive experiments in three real-world datasets have shown that our approach gives competitive results while requiring significantly less training time in comparison with other state-of-the-art methods.
\end{compactitem}







\section{Related works}





In this paper, we limit the scope to using correlation information between series to improve forecasting performance. Classical statistical methods, such as AutoRegressive Integrated Moving Average (ARIMA), are commonly used for univariate time series prediction, and its variant VAR is often used for multi-variate time series. In VAR, the relationship between series is represented by a matrix and therefore is linear. Deep learning-based methods such as LSTM \cite{lstm}, LSTNet \cite{lstnet} and TPA-LSTM \cite{tpa-lstm} can capture non-linear dependencies. These methods have a module that learns spatial information and encodes all of it on a single hidden state vector. This leads to the fact that the spatial information and the pairwise relationships between variables may not be fully exploited.


Constructing the relational graphs can help us better describe, analyze, and visualize the data, especially in domains such as transportation networks, social networks, and brain networks. In general, there are 2 main approaches, statistics-based such as probabilistic models, and deep learning-based, which often incorporate a graph generating module with a predicting model for downstream tasks in an end-to-end training framework. GraphLASSO \cite{10.1093/biostatistics/kxm045} is popular method in the first approach, it calculates an estimation of the precision matrix, which can represent the conditional correlations between each pair of variables. To better model a evolving graph, Time-Varying Graphical Lasso (TVGL) \cite{hallac2017network} uses temporal information to adjust the covariance matrix over time. On the another hand, several works in the second approach, such as \cite{MTGNN2020} and \cite{AGRCNbai2020}, learn an unique weighted adjacency matrix by computing the inner product between all pairs of nodes' the embedding vectors. To guarantee the necessary sparsity, \cite{MTGNN2020} selects top-k position in each row of the matrix, which is not flexible and dependent significantly on hyperparameter selection (i.e. $k$ is a hyperparameter). In a probabilistical point of view, one often assumes that the learnt dense matrix describes parameters of a high dimensional Bernoulli distribution of the correlation between series. LDS \cite{LDS} tackles the graph learning problem for non-temporal data by using a bi-level optimization routine and a straight-through gradient trick which, nonetheless, requires dense computation. The NRI approach \cite{NRI} learns a latent variable model predicting the interaction of physical objects via learning edge attributes of a fully-connected graph. GTS \cite{shang2021discrete} simplifies the NRI module by considering binary relationships only, and integrates graph inference in a spatiotemporal recurrent graph neural network \cite{li2017diffusion}. Both NRI and GTS exploit the categorical Gumbel trick \citep{GumbelJang2017, GumbelMaddison2017} to sample a discretized adjacency matrix while maintaining differentiability but still require dense computation. Most of aforementioned methods do not guarantee to produce a sparse graph, which is more intepretable and explanable in many real world datasets. Some of other works require pre-defined graph based on expert knowledge, such as RGSL \cite{RGSL}, to regularize sparsity of the structure. However, the expert knowledge is not avaiable in many cases and may not provide an accurate explanation for the complex relationship among data.

To address these above existing limitations in the field, our approach estimates the precision matrix with GraphLASSO, and uses it as parameters of a Gumbel distribution to sample the graph structure (i.e. adjacency matrix). Then, this structure is the input to a graph recurrent neural network module for time series forecasting as a downstream task. The decoupled training procedure in our proposal not only produces an effective graph structure but also has less training time in comparison with end-to-end training framework. Furthermore, we propose an unique loss function aiming to minimize the forecasting loss and construct the sparse graph structure of data simultaneously.
\section{Background}

In this study, we target on the Multi-variate Time-Series Forecasting (MTSF) problem. MTS can be denoted as a tensor $X \in \mathbb{R}^{N \times T}$, where $T$ is the number of timesteps, $N$ denote the number of series. At a time $t$, given a sequence $H$ time steps of historical observation, $X = \{X_{t-H+1}, X_{t-H+2},...,X_{t}\}$, we want to predict the future values of the next $F$ time steps, $\{X_{t+1}, X_{t+2},...,X_{t+F}\}$. Beside the observation, we can concatenate the auxiliary features $S_t$ to the input for improving forecasting performance.

Our approach models the relationships among entities (i.e. variables) by evolving graphs / networks. Since then, MTS data can be treated as spatial-temporal data. To accomplish the forecasting tasks, we will use STGNNs models. Here, we give some formal definition for the problem and algorithms.



\subsection{GraphLASSO} \label{sec:GraphLASSO}

GraphLASSO consider a sequence of multivariate observations in $\mathbb{R}^p$ sampled from a distribution $X  \sim N (0,  \Sigma(t))$. In static learning setting, the standard variation $\Sigma(t)$ is unique for all $t$. We estimate the precision matrix $\Theta  = \Sigma^{-1}$ with the constraint that $\Theta$ is symmetric positive-semidefinite by solving the following optimization problem:  
\begin{equation}\label{static_lasso}
\min_{\Theta \in S^P} \quad L(\Theta) =  - \log \text{det}(\Theta)  + \text{trace}(S \Theta) + \lambda \|\Theta \|_1
\end{equation}
where $S$ is the empirical covariance defined as $\frac{1}{n} X X^T$. When $S$ is invertible,  $L(\Theta)$ encourages $\Theta$ to converge to $S^{-1}$.
When $\Sigma(t)$ changes during time axis, we have an assumption that they are fixed in each given time interval $(t_i, t_{i-1})$.  Each time interval has at least 1 observation and we use these to estimate the corresponding precision matrix for each interval.
\cite{hallac2017network} provided a general formula of Time-Varying GraphLASSO (TVGL) problem:
\begin{equation}\label{eq:TVGL}
   \min_{\{\Theta_t\}_{t = 1}^T} \sum_{i=1}^{T} (-l_i (\Theta _i) + \lambda \|\Theta _i\|_1 )+ \beta \sum_{j=2}^{T} \psi (\Theta _j - \Theta_{j-1})
\end{equation}
where, $l_k (\Theta _k) = \log \text{det}(\Theta_k) - \text{trace}(S_k \Theta_k)$, $\beta \leq 0$ and  $\psi$ is a convex function, which gives penalty for the significant change between two $\Theta$s, see \cite{hallac2017network} for more details. 

\subsection{Graph Convolution Recurrent Network} \label{sec:GCRN}
Graph Convolutional Recurrent Network (GCRN) \cite{GCRN} is a deep learning model that can predict structured sequences of data. It is a generalization of classical recurrent neural networks (RNN) to data structured by an arbitrary graph. The main idea of GCRN is to merge different representations of the data provided by Graph Convolutional Network (GCN) layers and by recurrent layers.
GCN layers compute the spatial features of each node through a convolution operator that integrates information from each node's local neighbours.
According to \cite{Bruna2013, henaff2015deep, 10.5555/3157382.3157527}, the graph convolution operator can be well-approximated by 1st order Chebyshev polynomial expansion and can be generalized to high-dimensional GCN as:
\begin{equation}\label{eq:chevbyshev}
    G_{\star W} (X, A)= ( I + D^{-\frac{1}{2}} A D^{-\frac{1}{2}}) X W + W_b.
\end{equation}
The recurrent layers in GCRN are used to exploit dynamic patterns and capture the long-term dependencies in the sequential data. Basically, the hidden state at time step $t$ can be computed by a function of hidden at step $t-1$ and input at time $t$. There are several variations of recurrent can be applied here such as Long Short-Term Memoery (LSTM) \cite{lstm} or Gated Recurrent Unit (GRU) \cite{cho-etal-2014-learning}, etc.

\begin{figure*}[ht]
\centering
\includegraphics[width=0.8\textwidth]{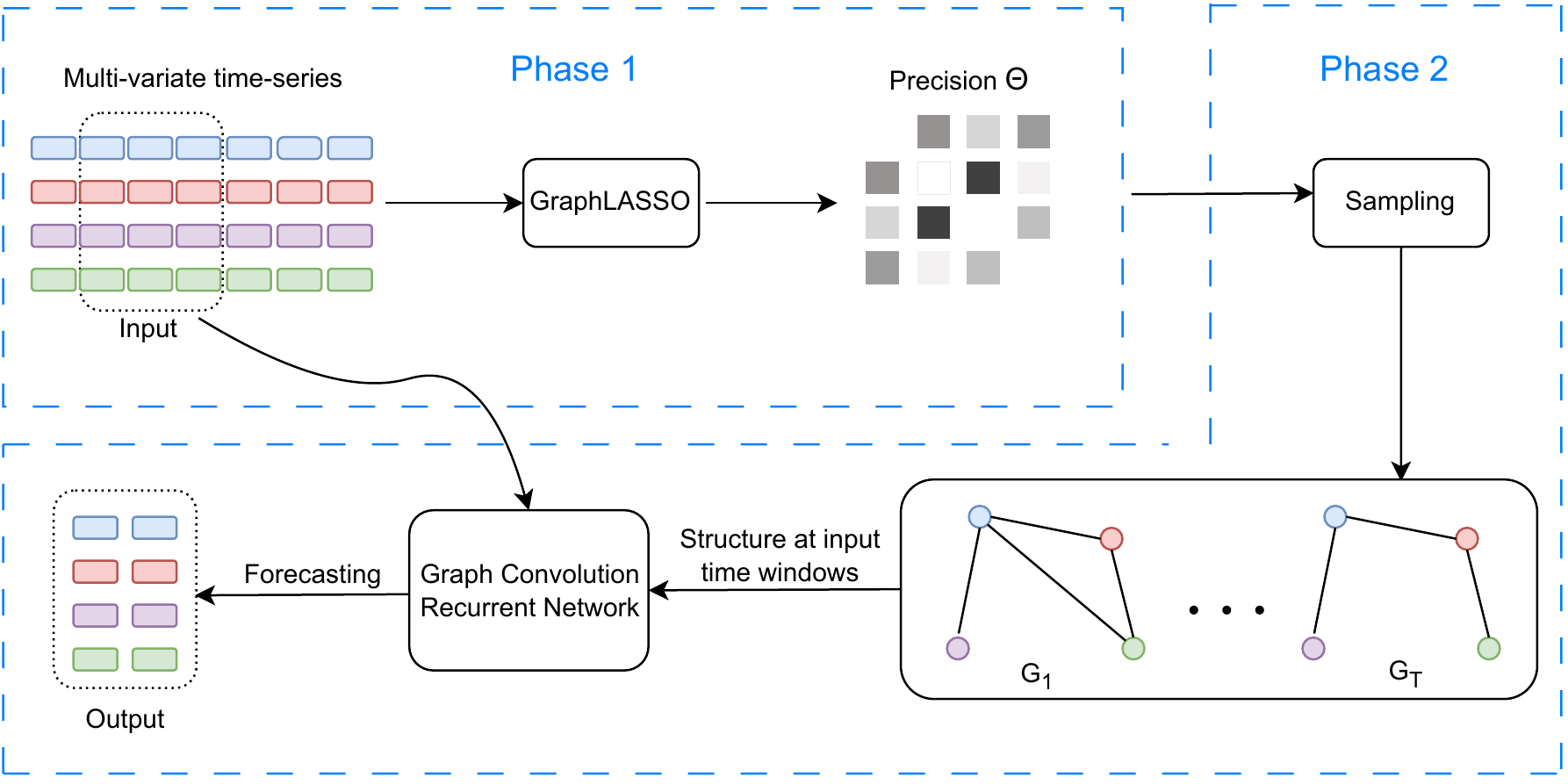}
\caption{\label{fig:decoupled}Decoupling architecture includes two phases. Phase 1 generates a sparse estimation of precision matrix. Depending on the case of a static structure or a time-varying structure, the output of Phase 1 is a matrix or a matrix sequence corresponding to the time intervals. Phase 2 samples the sparse graph structures based on these matrices and fit them into graph forecasting module.}
\end{figure*}

\section{Method}
We establish two phases sequentially that will be trained independently (see Figure \ref{fig:decoupled}). The first one is to estimate the precision matrices by using GraphLASSO. The second one is to forecast the future time series using graph convolution recurrent networks.

\subsection{Decoupled training architecture} 
Depending on whether there is a structural change in the network, we consider one of the two types of GraphLASSO problem presented in the section \ref{sec:GraphLASSO}. To align with Gaussian distribution assumption of GraphLASSO, we first detrend and normalize entire the time-series. The MTS dataset now is assumped that follow multivarite Gaussian distribution $ \mathcal{N} (0, \Sigma)$. The precision matrix $\Theta = \Sigma^{-1}$ represent the conditional dependence between any two pair of series. It means that $\Theta_{ij} = 0$ if and only if series $i$ is independent with series $j$ given other variables. 

In case of without assuming about the time-varying structures, we can use standard interior-point methods to solve \eqref{static_lasso}. We learn only one precision matrix $\Theta$ of entire dataset. 
In case of dynamically inferring graph structures, we can use Time-Varying Graphical Lasso (TVGL) algorithm to solve problem \eqref{eq:TVGL} (see \cite{hallac2017network}). This algorithm is based on the Alternating Direction Method of Multipliers (ADMM) \cite{ADMM}, a well-known approach for convex optimization that separates the problem into pieces that are easier to solve. The result is a sequence of $T$ precision matrices $\{\Theta_t\}_{t=1}^T$ corresponding to $T$ consecutive time intervals.  We choose $\psi(\Theta) = \|\Theta\|_F^2 = \sum_{i,j} \Theta_{ij}^2$ (i.e. $\|\cdot\|_F$ denotes the Frobenius norm) which can make smooth transitions of graph structure during time axis \cite{hallac2017network}.


%
Once the precision matrix $\Theta$ is estimated, we fit it into the graph forecasting module and sample the graph adjacency matrix $A$. We also consider $A_t$ to be a random variable with a parameterized distribution $P$ with the parameter $\Theta_k$, where $t$ is in time interval $(t_i, t_{i-1})$. Then the loss function of forecasting task will be rewritten as follows:
\begin{equation}
    \min \mathbb{E}_{A_t \sim P(\Theta)} \sum_t \textit{L} (f (X_{t+1:t+T}, A_t) , X_{t+T+1: t+T+F})
\end{equation}

In practice, we estimate the precision matrix on the rescaled dataset, so that $0 \leq \Theta_{ij} \leq 1$. We directly sample edge from node $i$ to $j$ by from the Bernoulli distribution ($\Theta_{ij}$), that means $A_{ij} = 1$ with probability $\Theta_{ij}$. Using a sampling technique instead of using a fixed threshold can help the model more flexible. Moreover, iterative sampling of the graph structures during training process can be seen as a dropout layer, which can decrease overfitting \cite{RGSL}. 







\subsection{Forecasting with graph neural network}
\label{sec:methodb}
In this part, we present the details of our forecasting module which is inspired by the GCRN architecture \cite{AGRCNbai2020, Dcrnn}. 
The module contains a Graph Convolution Module (GCM) which exploits spatial 
information and a Gate Recurrent Unit (GRU) network to capture temporal dependencies. The output of GCM with input $X$ and the corresponding adjacency matrix $A$ is denoted by $Z(X, A)$ as:
\begin{equation}\label{eq:graph_convolution}
Z(X,A) = G_{\star W}(X,A) = ( I + D^{-\frac{1}{2}} A D^{-\frac{1}{2}}) X W + W_b,
\end{equation}
where $I$ is identity matrix, $D$ is degree matrix of the graph, $W$ and $W_b$ are the learnable parameters. The updating mechanism of hidden state $h_t$ is given as follows:
\begin{equation} \label{gcgru}
\begin{split}
r_t & = \sigma(G_r(X_t\mathbin\Vert h_{t-1}, A), \\
z_t & = \sigma(G_z(X_t\mathbin\Vert h_{t-1}, A), \\
c_t & = \text{tanh}(G_c ( X_t \mathbin\Vert (r_t \odot h_{t-1}))),
\\
h_t & = z_t \odot h_{t-1} + (1-z_t) \odot c_t.
\end{split}
\end{equation}
Assume that we generate a sequence of graph adjacency matrices $\{A^k: k \in \{1, 2, \ldots, T\}\}$ corresponding to structures at $T$ time intervals. So, there are more than one graphs appear in a input-output sequence pair at some $t$. We tackle this issue by a simple aggregation strategy. Now, the graph convolution operator can be rewrited in the form:
\begin{equation}
\label{eq:aggre}
    G_{TV}(X) = \frac{1}{n}\sum_{A \in A^n } G(X,A),
\end{equation}
where $A^n$ is the set of $n$ graph structures that intersect with data sequence $X_{t-H+1: t+F}$. Finally, we use Mean Absolute Error (MAE) as  the loss function of forecasting module:
\begin{equation}
\label{eq:mae} 
L_f(\bm{W}) = \frac{1}{F}\|X_{t+1:t+F} - \hat{X}_{t+1:t+F}\|_1,
\end{equation}
where $\bm{W}$ denotes the set of all learnable parameters.






\section{Experiments}
In this section, extensive experiments will be conducted to
prove the efficacy and effectiveness of our proposed method, and the corresponding analysis will also be provided. Details about datasets and code can be found at \url{https://github.com/HySonLab/GraphLASSO}.

\subsection{Datasets}
We report results on two types of MTSF: (i) Small number of series including CA1, CA2, TX1, TX2, WI1, and WI2; and (ii) Large number of series including CA1-Food1, CA2-Food1, Electricity100, and Solar-energy-1h. Detail about these datasets can be founded in Appendix \ref{sec:Appendix}.

\subsection{Experiments setting}
\label{sec:exp_setting}
We compare the performance of our method with baseline methods as following: (i) HA: Historical Average, (ii) VAR: Vector Auto-Regressive model \cite{VAR}, (iii) LSTM: Long Short-Term Memory (LSTM) \cite{lstm}, (iv) GTS: An end-to-end framework for graph structure learning and GCRN forecasting \cite{shang2021discrete} and (v) $\text{GCRN}^*$: a GCRN model \cite{GCRN} that has the same architecture with our model but the input graph structure is fully connected graph (i.e. complete graph). We denote our model as $\text{Our}^{s}$ and $\text{Our}^{d}$, in which $\text{Our}^{s}$ includes a non-dynamic (i.e. static) graph learning module and $\text{Our}^{d}$ has a dynamic graph learning using time-varying GraphLASSO module. The architecture of the corresponding GCRN forecasting modules were presented in \ref{sec:methodb}.  

In the benchmark on M5 datasets, we use 14 historical timesteps to predict 7 future timesteps. In electricity and solar-energy, we use 168 input steps and predict 1 output step. Details about other parameters and training for all models can be founded in Appendix \ref{sec:appendix2}.

\subsection{Results}
\label{sec:result}
\begin{table}[th]
\centering
\resizebox{0.9\columnwidth}{!}{%
\begin{tabular}{ccccccc}
\toprule
& HA     & VAR    & LSTM            & GTS    & GCRN   & \textbf{Ours} \\ 
\midrule
CA1 & 0.1180 & 0.1063 & 0.1026          & 0.0933 & 0.0956 & \textbf{0.0902} \\
CA2 & 0.1311 & 0.1277 & 0.0692          & 0.0698 & 0.0713 & \textbf{0.0691} \\
TX1 & 0.1038 & 0.0620 & 0.0659          & 0.0665 & 0.0671 & \textbf{0.0616} \\
TX2 & 0.0854 & 0.0625 & \textbf{0.0619} & 0.0623 & 0.0684 & 0.0621          \\
WI1 & 0.1608 & 0.0768 & 0.0787          & 0.0723 & 0.0730 & \textbf{0.0715} \\
WI2 & 0.1488 & 0.0881 & \textbf{0.0732} & 0.0801 & 0.0843 & 0.0762          \\
\bottomrule
\end{tabular}%
}
\caption{Mean absolute error of M5 store-level datatset}
\label{tab:small}
\end{table}

\begin{table*}[ht]
\centering
\resizebox{0.9\textwidth}{!}
{
\begin{tabular}{cccccccccccc}
\toprule

&
  \multicolumn{1}{c}{HA} &
  \multicolumn{1}{c}{VAR} &
  \multicolumn{1}{c}{LSTM} &
  \multicolumn{2}{c}{GTS} &
  \multicolumn{2}{c}{GCRN} &
  \multicolumn{2}{c}{$\text{Our}^s$} &
  \multicolumn{2}{c}{$\text{Our}^d$} \\
 &
  \multicolumn{1}{c}{MAE} &
  \multicolumn{1}{c}{MAE} &
  \multicolumn{1}{c}{MAE} &
  \multicolumn{1}{c}{MAE} &
  \multicolumn{1}{c}{t-time} &
  \multicolumn{1}{c}{MAE} &
  \multicolumn{1}{c}{t-time} &
  \multicolumn{1}{c}{MAE} &
  \multicolumn{1}{c}{t-time} &
  \multicolumn{1}{c}{MAE} &
  \multicolumn{1}{c}{t-time} \\
  \midrule

CA1\_Food1   & 1.9869   & 2.2169   & 1.4045  & \textbf{1.3601}  & 70s  & 1.5108  & 46s  & 1.4019          & \textit{37s}  & 1.3916          & \textit{40s}  \\
TX1\_Food1   & 1.7883   & 2.3486   & 1.6067  & 1.3741           & 75s  & 1.4061  & 43s  & \textbf{1.3629} & \textit{34s}  & 1.3777          & \textit{38s}  \\
WI1\_Food1   & 1.8058   & 2.3056   & 1.4335  & \textbf{1.0818}  & 72s  & 1.1238  & 45s  & 1.0843          & \textit{35s}  & 1.0964          & \textit{38s}  \\
Electricity100  & 231.3055 & 101.2598 & 91.6094 & \textbf{70.4458} & 390s & 76.5002 & 310s & 75.8671         & \textit{215s} & 72.6113         & \textit{230s} \\
Solar-energy-1h & 8.2004   & 4.4045   & 2.5334  & 1.8263           & 125s & 2.0925  & 95s  & 1.8406          & \textit{75s}  & \textbf{1.1802} & \textit{80s} \\
\bottomrule

\end{tabular}%
}
\caption{Forecasting error and time training evaluated on 5 MTS datasets which have more than 100 time series.}
\label{tab:large}
\end{table*}

We first report the MAE results of the next 7th step prediction on datasets that have small number of series in Table \ref{tab:small}.
The historical average method has the most misleading prediction among the models. 
Deep learning models have better results than VAR on all 6 datasets, which indicates that representing inter-relationships using non-linear models will be more effective than the linear one. LSTM gives the best performance in 2 stores, TX2 and WI2, while our model is best in the remaining 4 stores. Among the models using GNNs, our model gives better results than GTS and $\text{GCRN}^*$. This shows that the graph produced by GraphLasso can provide more effective spatial information for the GNN forecasting model. Moreover, our model can control the sparsity of the generated graphs and it can give a reasonable explanation for the relationship between series. 

We report the results of comparing our models and baselines on MTS datasets with a larger number of series as mentioned above including MAE error and training time denoted \textit{t-time} in Table \ref{tab:large}. 
Our methods have competitive results against baselines. Model $\text{Our}^s$, which uses static GraphLASSO, has best MAE loss in $\text{TX1}\_\text{Food1}$ dataset and our dynamic model archives the best result in Solar-energy-1h dataset. In particular, there was a shift distribution in Solar-energy dataset because it depends on sunlight intensity during the year. So it may contain a complex time-varying correlation and the relationships between nodes change over time. Model $\text{Our}^d$, which uses time-varying GraphLASSO for graph learning, can capture this change well and makes better predictions than the others.

In the remaining 3 datasets, our models have results that are not too much worse than the best baseline but significantly reduce the training time. 
Our model computes the relationship graph based on the statistical estimation of the precision matrix, thus reducing both the computational complexity and the number of parameters to be learned by the graph learning module compared to an end-to-end approach.
Another advantage of our models is the use of sparse graphs, which can be implemented in the form of sparse matrices, thus reducing the computation time of graph convolution operators.
In summary, our method can capture accurately both spatial information and temporal information in the correlated MTS data and significantly reduce the training time in comparison with other GNN-based methods.

\section{Conclusion}

In this paper, we have proposed a graph forecasting method consisting of two separate training modules to solve the MTSF problem. The first module learns the sparse graph structures that describe the correlation between series in a data-driven manner. We use GraphLASSO to directly estimate the precision matrix of data to construct the graph structures and control their sparsity. We consider two types of learning graph structures: the relationship between series is unchanged / static (i.e. independent of time) or dependent on each specific time interval (i.e. time-varying). In the static case, only one graph is generated. In the time-varying case, we construct a sequence of graphs that describe the pairwise relationships between series.
The second module is built from GCRN that takes both the data and the graph structures generated from the first module as input to learn a forecasting model. Taking advantage of the sparsity, we can reduce the training time of both the graph generating module and the time-series forecasting module.
Our experiments on three datasets can be divided into two cases: small number of series and large number of series. In both cases, our models not only show a competitive performance but also reduce training time by $\sim 40\%$ in comparison with other forecasting baselines. 
In future works, we aim to incorporate the end-to-end learning and experts' knowledge graphs to making more meaningful explanation and improve forecasting performance.

\bibliographystyle{ieeetr}
\bibliography{main}

\begin{thebibliography}{10}

\bibitem{GCRNtraffic}
Z.~Cui, K.~Henrickson, R.~Ke, and Y.~Wang, ``Traffic graph convolutional
  recurrent neural network: A deep learning framework for network-scale traffic
  learning and forecasting,'' {\em IEEE Transactions on Intelligent
  Transportation Systems}, vol.~21, no.~11, pp.~4883--4894, 2019.

\bibitem{Dcrnn}
Y.~Li, R.~Yu, C.~Shahabi, and Y.~Liu, ``Diffusion convolutional recurrent
  neural network: Data-driven traffic forecasting,'' in {\em 6th International
  Conference on Learning Representations, {ICLR} 2018, Vancouver, BC, Canada,
  April 30 - May 3, 2018, Conference Track Proceedings}, OpenReview.net, 2018.

\bibitem{pmlr-v184-hy22a}
T.~S. Hy, V.~B. Nguyen, L.~Tran-Thanh, and R.~Kondor, ``Temporal
  multiresolution graph neural networks for epidemic prediction,'' in {\em
  Proceedings of the 1st Workshop on Healthcare AI and COVID-19, ICML 2022}
  (P.~Xu, T.~Zhu, P.~Zhu, D.~A. Clifton, D.~Belgrave, and Y.~Zhang, eds.),
  vol.~184 of {\em Proceedings of Machine Learning Research}, pp.~21--32, PMLR,
  22 Jul 2022.

\bibitem{retailsales}
S.~Liao, J.~Yin, and W.~Rao, ``Towards accurate retail demand forecasting using
  deep neural networks,'' in {\em Database Systems for Advanced Applications}
  (Y.~Nah, B.~Cui, S.-W. Lee, J.~X. Yu, Y.-S. Moon, and S.~E. Whang, eds.),
  (Cham), pp.~711--723, Springer International Publishing, 2020.

\bibitem{finance}
A.~Patton, ``Chapter 16 - copula methods for forecasting multivariate time
  series,'' in {\em Handbook of Economic Forecasting} (G.~Elliott and
  A.~Timmermann, eds.), vol.~2 of {\em Handbook of Economic Forecasting},
  pp.~899--960, Elsevier, 2013.

\bibitem{VAR}
J.~Stock and M.~Watson, ``Vector autoregressions,'' {\em Journal of Economic
  Perspectives}, vol.~15, no.~4, p.~101 {\textendash} 116, 2001.

\bibitem{VARMA}
H.~Spliid, ``A fast estimation method for the vector autoregressive moving
  average model with exogenous variables,'' {\em Journal of the American
  Statistical Association}, vol.~78, no.~384, pp.~843--849, 1983.

\bibitem{logformer}
S.~Li, X.~Jin, Y.~Xuan, X.~Zhou, W.~Chen, Y.-X. Wang, and X.~Yan, ``Enhancing
  the locality and breaking the memory bottleneck of transformer on time series
  forecasting,'' {\em Advances in neural information processing systems},
  vol.~32, 2019.

\bibitem{crossformer}
Y.~Zhang and J.~Yan, ``Crossformer: Transformer utilizing cross-dimension
  dependency for multivariate time series forecasting,'' in {\em The Eleventh
  International Conference on Learning Representations}, 2022.

\bibitem{STGCN}
B.~Yu, H.~Yin, and Z.~Zhu, ``Spatio-temporal graph convolutional networks: A
  deep learning framework for traffic forecasting,'' in {\em Proceedings of the
  27th International Joint Conference on Artificial Intelligence}, IJCAI'18,
  p.~3634–3640, AAAI Press, 2018.

\bibitem{STGNN}
D.~Cao, Y.~Wang, J.~Duan, C.~Zhang, X.~Zhu, C.~Huang, Y.~Tong, B.~Xu, J.~Bai,
  J.~Tong, and Q.~Zhang, ``Spectral temporal graph neural network for
  multivariate time-series forecasting,'' in {\em Proceedings of the 34th
  International Conference on Neural Information Processing Systems}, NIPS'20,
  (Red Hook, NY, USA), Curran Associates Inc., 2020.

\bibitem{AGRCNbai2020}
L.~Bai, L.~Yao, C.~Li, X.~Wang, and C.~Wang, ``Adaptive graph convolutional
  recurrent network for traffic forecasting,'' in {\em Proceedings of the 34th
  International Conference on Neural Information Processing Systems}, NIPS'20,
  (Red Hook, NY, USA), Curran Associates Inc., 2020.

\bibitem{li2017diffusion}
Y.~Li, R.~Yu, C.~Shahabi, and Y.~Liu, ``Diffusion convolutional recurrent
  neural network: Data-driven traffic forecasting,'' {\em arXiv preprint
  arXiv:1707.01926}, 2017.

\bibitem{MTGNN2020}
Z.~Wu, S.~Pan, G.~Long, J.~Jiang, X.~Chang, and C.~Zhang, ``Connecting the
  dots: Multivariate time series forecasting with graph neural networks,'' in
  {\em Proceedings of the 26th ACM SIGKDD International Conference on Knowledge
  Discovery \& Data Mining}, KDD '20, (New York, NY, USA), p.~753–763,
  Association for Computing Machinery, 2020.

\bibitem{4700287}
F.~Scarselli, M.~Gori, A.~C. Tsoi, M.~Hagenbuchner, and G.~Monfardini, ``The
  graph neural network model,'' {\em IEEE Transactions on Neural Networks},
  vol.~20, no.~1, pp.~61--80, 2009.

\bibitem{kipf2017semisupervised}
T.~N. Kipf and M.~Welling, ``Semi-supervised classification with graph
  convolutional networks,'' in {\em International Conference on Learning
  Representations}, 2017.

\bibitem{shang2021discrete}
C.~Shang and J.~Chen, ``Discrete graph structure learning for forecasting
  multiple time series,'' in {\em Proceedings of International Conference on
  Learning Representations}, 2021.

\bibitem{10.1093/biostatistics/kxm045}
J.~Friedman, T.~Hastie, and R.~Tibshirani, ``{Sparse inverse covariance
  estimation with the graphical lasso},'' {\em Biostatistics}, vol.~9,
  pp.~432--441, 12 2007.

\bibitem{10.1093/biomet/asm018}
M.~Yuan and Y.~Lin, ``{Model selection and estimation in the Gaussian graphical
  model},'' {\em Biometrika}, vol.~94, pp.~19--35, 03 2007.

\bibitem{lstm}
S.~Hochreiter and J.~Schmidhuber, ``Long short-term memory,'' {\em Neural
  Comput.}, vol.~9, p.~1735–1780, nov 1997.

\bibitem{lstnet}
G.~Lai, W.-C. Chang, Y.~Yang, and H.~Liu, ``Modeling long-and short-term
  temporal patterns with deep neural networks,'' in {\em The 41st international
  ACM SIGIR conference on research \& development in information retrieval},
  pp.~95--104, 2018.

\bibitem{tpa-lstm}
S.-Y. Shih, F.-K. Sun, and H.-y. Lee, ``Temporal pattern attention for
  multivariate time series forecasting,'' {\em Machine Learning}, vol.~108,
  pp.~1421--1441, 2019.

\bibitem{hallac2017network}
D.~Hallac, Y.~Park, S.~Boyd, and J.~Leskovec, ``Network inference via the
  time-varying graphical lasso,'' in {\em Proceedings of the 23rd ACM SIGKDD
  International Conference on Knowledge Discovery and Data Mining},
  pp.~205--213, 2017.

\bibitem{LDS}
L.~Franceschi, M.~Niepert, M.~Pontil, and X.~He, ``Learning discrete structures
  for graph neural networks,'' in {\em Proceedings of the 36th International
  Conference on Machine Learning} (K.~Chaudhuri and R.~Salakhutdinov, eds.),
  vol.~97 of {\em Proceedings of Machine Learning Research}, pp.~1972--1982,
  PMLR, 09--15 Jun 2019.

\bibitem{NRI}
T.~Kipf, E.~Fetaya, K.-C. Wang, M.~Welling, and R.~Zemel, ``Neural relational
  inference for interacting systems,'' in {\em Proceedings of the 35th
  International Conference on Machine Learning} (J.~Dy and A.~Krause, eds.),
  vol.~80 of {\em Proceedings of Machine Learning Research}, pp.~2688--2697,
  PMLR, 10--15 Jul 2018.

\bibitem{GumbelJang2017}
E.~Jang, S.~Gu, and B.~Poole, ``Categorical reparameterization with
  gumbel-softmax,'' in {\em 5th International Conference on Learning
  Representations, {ICLR} 2017, Toulon, France, April 24-26, 2017, Conference
  Track Proceedings}, OpenReview.net, 2017.

\bibitem{GumbelMaddison2017}
C.~J. Maddison, A.~Mnih, and Y.~W. Teh, ``The concrete distribution: {A}
  continuous relaxation of discrete random variables,'' in {\em 5th
  International Conference on Learning Representations, {ICLR} 2017, Toulon,
  France, April 24-26, 2017, Conference Track Proceedings}, OpenReview.net,
  2017.

\bibitem{RGSL}
H.~Yu, T.~Li, W.~Yu, J.~Li, Y.~Huang, L.~Wang, and A.~Liu, ``Regularized graph
  structure learning with semantic knowledge for multi-variates time-series
  forecasting,'' in {\em Proceedings of the Thirty-First International Joint
  Conference on Artificial Intelligence, {IJCAI-22}} (L.~D. Raedt, ed.),
  pp.~2362--2368, International Joint Conferences on Artificial Intelligence
  Organization, 7 2022.
\newblock Main Track.

\bibitem{GCRN}
Y.~Seo, M.~Defferrard, P.~Vandergheynst, and X.~Bresson, ``Structured sequence
  modeling with graph convolutional recurrent networks,'' in {\em Neural
  Information Processing: 25th International Conference, ICONIP 2018, Siem
  Reap, Cambodia, December 13-16, 2018, Proceedings, Part I 25}, pp.~362--373,
  Springer, 2018.

\bibitem{Bruna2013}
J.~Bruna, W.~Zaremba, A.~D. Szlam, and Y.~LeCun, ``Spectral networks and
  locally connected networks on graphs,'' {\em CoRR}, vol.~abs/1312.6203, 2013.

\bibitem{henaff2015deep}
M.~Henaff, J.~Bruna, and Y.~LeCun, ``Deep convolutional networks on
  graph-structured data,'' {\em arXiv preprint arXiv:1506.05163}, 2015.

\bibitem{10.5555/3157382.3157527}
M.~Defferrard, X.~Bresson, and P.~Vandergheynst, ``Convolutional neural
  networks on graphs with fast localized spectral filtering,'' in {\em
  Proceedings of the 30th International Conference on Neural Information
  Processing Systems}, NIPS'16, (Red Hook, NY, USA), p.~3844–3852, Curran
  Associates Inc., 2016.

\bibitem{cho-etal-2014-learning}
K.~Cho, B.~van Merri{\"e}nboer, C.~Gulcehre, D.~Bahdanau, F.~Bougares,
  H.~Schwenk, and Y.~Bengio, ``Learning phrase representations using {RNN}
  encoder{--}decoder for statistical machine translation,'' in {\em Proceedings
  of the 2014 Conference on Empirical Methods in Natural Language Processing
  ({EMNLP})}, (Doha, Qatar), pp.~1724--1734, Association for Computational
  Linguistics, Oct. 2014.

\bibitem{ADMM}
S.~P. Boyd, N.~Parikh, E.~Chu, B.~Peleato, and J.~Eckstein, ``Distributed
  optimization and statistical learning via the alternating direction method of
  multipliers,'' {\em Found. Trends Mach. Learn.}, vol.~3, no.~1, pp.~1--122,
  2011.

\bibitem{M5data}
S.~Makridakis, E.~Spiliotis, and V.~Assimakopoulos, ``M5 accuracy competition:
  Results, findings, and conclusions,'' {\em International Journal of
  Forecasting}, vol.~38, no.~4, pp.~1346--1364, 2022.

\bibitem{GGLASSO}
F.~Schaipp, O.~Vlasovets, and C.~L. Müller, ``Gglasso - a python package for
  general graphical lasso computation,'' {\em Journal of Open Source Software},
  vol.~6, no.~68, p.~3865, 2021.

\bibitem{kingma2014adam}
D.~P. Kingma and J.~Ba, ``Adam: A method for stochastic optimization,'' {\em
  arXiv preprint arXiv:1412.6980}, 2014.

\end{thebibliography}

\clearpage
\appendices    
\section{Dataset and model parameter}
\subsection{Dataset}
\label{sec:Appendix}
We used M5 retail \cite{M5data} and its sub-dataset in our experiments.
The M5 is a retail sales forecasting dataset that uses real-life, hierarchically structured sales data with intermittent characteristics.
It collects retail data at 10 stores in 3 states in the United States including California (CA), Texas (TX) and Wisconsin (WI). Each store contains 3,049 items and they are divided into 7 departments (e.g., Food1, Food2, Household1, etc.).  All the series were daily recorded in a period of 1,943 days. The information about items and departments is encoded, hence we cannot provide an expert knowledge graph for the model. In this paper, we used 6 store-level series as subsets of original M5, which are CA1, CA2, TX1, TX2, WI1, and WI2. This store-level data is then be aggregated by its department, so the dataset has 7 series and 1,943 timesteps. We created item-level subset grouped by the department in each store because the items in same department may have important dependencies. Data in some departments contains many zero values which are related to disruption in the selling process of items. To reduce data processing and give balance comparision with baselines, we only report result on Food1 department which has 216 items. 

We also evaluated our model on other commonly used MTS datasets which are: Electricity, Solar-energy \cite{MTGNN2020}. The origin Electricity dataset contains 321 series represent the electricity consumption of clients from 2012 to 2014 and the length of series is 26,304. We only used first 100 series in this dataset in our experiments. Solar-energy dataset includes solar power output
collected from 137 sensors in Alabama state in 2007. The timestep in Solar-energy is aggregated to 1 hour, and the number of samples is 8,760. We denoted the two new datasets are Electricity100 and Solar-energy-1h.

We present some statstic detail about these datasets in \ref{tab:data}. 
\begin{table}[hb]
\centering
\resizebox{0.8\columnwidth}{!}{%
\begin{tabular}{cccc}
\toprule
Dataset         & Mean   & Std    & Data range   \\
\midrule
Electricity100  & 650.14 & 283.07 & 0 $-$ 15443 \\
Solar-energy-1h & 6.35   & 9.10   & 0 $-$ 86.85 \\
CA1\_Food1      & 1.52   & 2.86   & 0 $-$ 92    \\
TX1\_Food1      & 1.06   & 2.42   & 0 $-$ 108   \\
WI1\_Food1      & 1.44   & 2.83   & 0 $-$ 166   \\
CA1             & 634.28 & 656.17 & 0 $-$ 4042  \\
TX1             & 453.41 & 452.41 & 0 $-$ 2919  \\
WI1             & 511.36 & 493.51 & 0 $-$ 3203  \\
\bottomrule
\end{tabular}%
}
\
\caption{Statistics of MTS datasets used in experiments.}
\label{tab:data}
\end{table}

\subsection{Data preprocessing}
Because the M5 retail data at item-level is intermittent and erratic, we first used $\log(1+x)$ transformation on input sale data. For the department-level M5 data, we used min-max scaling transformation in each department to keep the data range in (0,1).
We applied normalizing transformation for all datasets before estimating the precision matrix:
\begin{equation}
    \hat{X} = \dfrac{X- \text{mean}(X)}{\text{std}(X)}
\end{equation}
We also used day-in-week and hour-in-day as auxiliary temporal data to improve forecasting performance. We generated sequence-to-sequence type data for deep learning models. The train/validate/test ratio was separated as ratio $70\%/10\%/20\%$.

\subsection{Evaluate metrics}
We used Mean Absolute Error (MAE) to measure forecasting performance. We evaluated forecasting results at $F-th$ timestep in the future, we have the MAE error between ground truth and the prediction of all series as:
\begin{equation}
    \dfrac{1}{D} \sum_{j=1}^D |X_{t+F,j}-\hat{X}_{t+F,j}|
\end{equation}
where $D$ is the number of series.
We also reported the training time on larger number of series datasets. Total training time including Phase 1 and Phase 2 of our models were divided by the number of epochs. We compare them with the average training time of each epoch of other baselines. The unit measure is second. All the experiments were  performed on a same computer with GPU NVIDIA GTX 1650Ti.

\subsection{Implementation details}
\label{sec:appendix2}
The baselines have the detail of implementation as follow:
\begin{itemize}
    \item HA: We use the historical timestamp value of the past 24 steps in Electricity100 and Solar-energy-1h datasets to predict 6 steps in the future and the past 14 steps in the past and predict 7 future steps in the remaining datasets. All the predictions in the horizon are invariant.
    \item VAR: We implement the VAR model based on the python package \textit{statsmodel}. The lag parameter was selected as 7 for all M5 sub-datasets and 12 for the remaining.
    \item LSTM: We implement an encoder-decoder model which two layers of LSTM with Tensorflow. Each layer has 32 and 64 hidden units for M5 datasets and the remaining, respectively. A fully-connected layer was added to predict the output.
    \item $\text{GCRN}^*$: We set the maximum step of randoms walks to 2, the initial learning rate to 0.01, drop out rate to 0.1 for all datasets, the GRU layer has 32 and 64 hidden units for M5 datasets and the remaining, respectively.
    \item GTS: We used the same architecture for the graph learning module with 2 layers of CNN and 2 layers of MLP as in \cite{shang2021discrete} in the M5 sub-datasets. In the other datasets, we used 3 layers of CNN and 2 MLP layers. The batch size was selected by 32 for all the datasets. Learning rate and hidden units were chosen the same with the $\text{GCRN}^*$ model. Max diffusion step was selected by 2 and the number of RNN layers was 1.
\end{itemize}
In our two models $\text{Our}^s$ and $\text{Our}^d$, the precision matrices were estimated by using the GGLASSO package \cite{GGLASSO}. The $\lambda$ and $\beta$ parameters in  Equation \ref{eq:TVGL} were 0.1 and 0.05, respectively. For all the deep learning models including ours, we optimize with the Adam optimizer \cite{kingma2014adam} for 100 epochs and use an early stopping strategy with the patience of 15 epochs by monitoring the loss in the validation set. After hyperparameter searching, we chose the initial learning rate as 0.005 for all datasets. Other parameters were the same with the $\text{GCRN}^*$ model \cite{GCRN}.
\end{document}